\documentclass{llncs}
\usepackage{makeidx}  
\usepackage{graphicx}
\usepackage{wrapfig}
\usepackage{subfig}
\usepackage{amsmath}
\usepackage{multirow}
\usepackage{array}
\usepackage{bbm}
\usepackage{textcomp}
\usepackage{tabularx}

\begin{document}
\mainmatter              
\title{Cardiac Motion Scoring with Segment- and Subject-level Non-Local Modeling}

\titlerunning{Cardiac Motion Scoring with Non-local modeling}  

\author{Wufeng Xue\inst{1,2} \and Gary Brahm\inst{1} \and Stephanie Leung\inst{1}\and \\ Ogla Shmuilovich\inst{1} \and Shuo Li*\inst{1}}
\authorrunning{W. Xue, G. Brahm, S. Leung, et al.} 


\institute{Department of Medical Imaing, Western University, ON, Canada \email{slishuo@gmail.com}
\and National-Reginal Key Technology Engineering Laboratory for Medical Ultrasound, School of Biomedical Engineering, Health Science Center, Shenzhen University, China
}

\maketitle              

\begin{abstract}
Motion scoring of cardiac myocardium is of paramount importance for early detection and diagnosis of various cardiac disease. 
It aims at identifying regional wall motions into one of the four types including normal, hypokinetic, akinetic, and dyskinetic, and is extremely challenging due to the complex myocardium deformation and subtle inter-class difference of motion patterns. All existing work on automated motion analysis are focused on binary abnormality detection to avoid the much more demanding motion scoring, which is urgently required in real clinical practice yet has never been investigated before. In this work, we propose Cardiac-MOS, the first powerful method for cardiac motion scoring from cardiac MR sequences based on deep convolution neural network. Due to the locality of convolution, the relationship between distant non-local responses of the feature map cannot be explored, which is closely related to motion difference between segments. In Cardiac-MOS, such non-local relationship is modeled with non-local neural network within each segment and across all segments of one subject, i.e., segment- and subject-level non-local modeling, and lead to obvious performance improvement. Besides, Cardiac-MOS can effectively extract motion information from MR sequences of various lengths by interpolating the convolution kernel along the temporal dimension, therefore can be applied to MR sequences of multiple sources.       
Experiments on 1440 myocardium segments of 90 subjects from short axis MR sequences of multiple lengths prove that Cardiac-MOS achieves reliable performance, with correlation of 0.926 for motion score index estimation and accuracy of 77.4\% for motion scoring. Cardiac-MOS also outperforms all existing work for binary abnormality detection. As the first automatic motion scoring solution, Cardiac-MOS demonstrates great potential in future clinical application. 

\end{abstract}

\section{Introduction}

Cardiac motion scoring is essential for clinical diagnosis of various cardiac disease~\cite{Lang2015Recommendations}, including coronary artery disease, congestive heart failure, stress-induced cardiomyopathy, myocarditis, stroke, amongst others. However, cardiac wall motion is a complex deformation procedure with regional wall thickening, circumferential shortening and longitudinal ventricular shortening.  It varies a lot in the presence of different types of pathology. In routine clinical practice, motion scoring is often conducted by labor-intensive visual inspection of the dynamic cardiac sequences of MR or Echocardiograms for each segment of left ventricle, following the scoring system of~\cite{Lang2015Recommendations}: (1) normal, (2) hypokinetic, (3) akinetic, and (4) dyskinetic. The results obtained in this way are characterized by large inter-rater variability and low reproducibility~\cite{Paetsch2006determination} due to the complex regional motion patterns and the subtle motion difference between segments.

However, automated motion scoring from cardiac MR images has never been investigated despite its clinical significance. Existing work only focused on binary motion abnormality detection\cite{Afshin2014Regional,Afshin2011assessment,Leung2007localized,Lu2009Pattern,Mantilla2015Classification,Punithakumar2012regional,Punithakumar2013regional,Punithakumar2010regional,Punithakumar2010detection,Qian2008identifying,Qian2011identifying,Suinesiaputra2009automated}, which alleviates a lot the difficulty of differentiating the subtle difference among various motion patterns.
In summary, these methods follow a pipeline of: 1) \emph{myocardium segments localization}, by manually or semi-automatically delineating the contours of myocardium~\cite{Afshin2014Regional,Afshin2011assessment,Leung2007localized,Lu2009Pattern,Suinesiaputra2009automated,Punithakumar2012regional,Punithakumar2013regional,Punithakumar2010regional,Punithakumar2010detection}, 2) handcrafted \emph{motion information extraction}, including spatial-temporal profiles~\cite{Qian2008identifying,Qian2011identifying}, inter-frame correlations~\cite{Afshin2011assessment,Afshin2014Regional,Lu2009Pattern}, or parameter distribution~\cite{Punithakumar2012regional,Punithakumar2013regional,Punithakumar2010regional,Punithakumar2010detection}, and then 3) \emph{motion classification}.

\begin{table}[!t]
	\caption{Existing methods on \emph{binary} abnormality detection only obtained good performance for small dataset with LOO experiment setting.}
	\label{table_methods_review}
	\centering
	\begin{tabular}{l|ccccc}
		\hline
		Methods& $\#$ of Subject&Modality& Training/Test&Accuracy(\%)&Kappa value\\
		\hline
		\cite{Punithakumar2010regional}&30&SAX MR&LOO&90.8&0.738\\
		\cite{Punithakumar2012regional}&30&SAX, 2C, 3C, 4C MR&LOO&91.9&0.738\\
		\cite{Punithakumar2013regional}&58&SAX MR&LOO&87.1&0.73\\
		\cite{Qian2011identifying,Qian2008identifying}&22&SAX Tagged MR&LOO&87.8&-\\
		\cite{Lu2009Pattern}&17& SAX Basal MR&No split&86.3&0.693\\
		\cite{Afshin2014Regional}&58&SAX MR&3-fold CV&86&0.73\\
		\cite{Suinesiaputra2009automated}&89&SAX MR&45/44&65.9&-\\
		\cite{Leung2007localized}&129&2C and 4C Echo&65/64&76.5&-\\		
		\hline
	\end{tabular}
	\footnotesize{SAX: short axis view; 2C: two chamber view; 3C: three chamber view; 4C: four chamber view; LOO: leave-one-subject-out; CV: cross validation}
\end{table}

While these methods achieved promising accuracy for abnormality detection, they cannot be applied to the task of motion scoring because 1) only heuristic handcrafted feature is not sufficient to capture the complex regional motion information; and 2) absence of dependency modeling between local features  of all segments cannot capture the subtle motion difference between segments. Besides, they still suffer from the following limitations: 1) delineation of myocardium border is required, either manually or semi-automatically, introducing inconvenience in clinical practice; 2) multiple classifiers are trained for different segments~\cite{Leung2007localized,Lu2009Pattern,Afshin2011assessment,Afshin2014Regional} or slices~\cite{Punithakumar2012regional,Punithakumar2013regional,Punithakumar2010regional,Suinesiaputra2009automated}, complicating their practical application; 3) only achieves good performance on small dataset (as shown in Table.~\ref{table_methods_review}), where the diversity of cardiac motion pattern is limited.

In this paper, we take advantage of the deep convolution network and non-local modeling~\cite{Wang2017nonlocal}, and propose the first reliable solution, Cardiac-MOS for Cardiac MOtion Scoring. While the convolution network extracts discriminative \emph{local} features, non-local modeling allows distant \emph{non-local} responses to contribute to the response at a location, therefore capturing the subtle relative changes of local motion features. In Cardiac-MOS, we 1) design a kernel-interpolation based convolution (conv-KI) layer, which captures motion information by convolution along the temporal dimension, and reuses the kernel by interpolation for sequences of various lengths, and 2) introduce segment- and subject-level non-local modeling to explore the long-range relationships within each segment and across all 16 segments of one subject. With non-local modeling, Cardiac-MOS is capable of extracting the subtle motion difference between segments and thus delivering accurate motion scoring results.

\begin{figure}[!t]
	\centering
	\includegraphics[width=10cm]{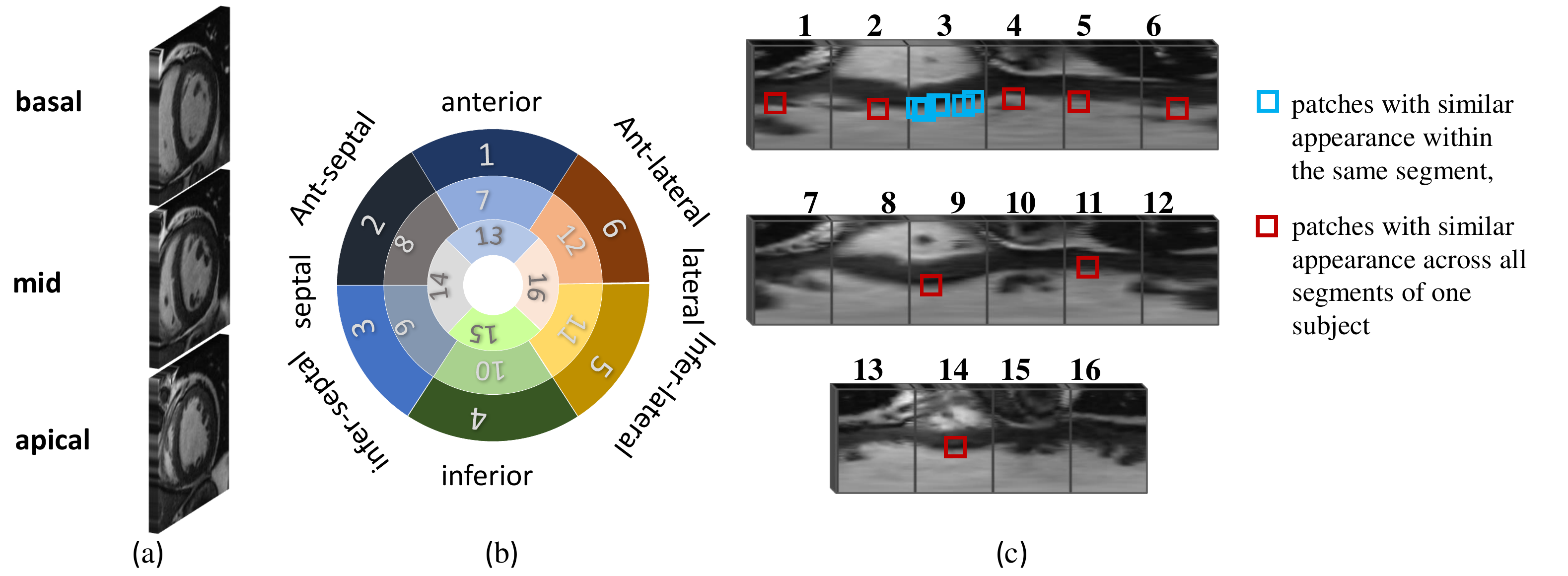}
	\caption{Myocardium segment re-sampling. The 16 segments obtained from short axis slices (a) according to the AHA 17-segment model~\cite{Lang2015Recommendations} of left ventricle (b) are converted into the polar coordinates and re-sampled to rectangular sequences of the same size (c), so that the motion information can be more easily modeled. Appearance similarity between patches within each segment (blue) and patches across all segments (red) underlies our non-local modeling.}
	\label{fig_16seg_polar}
\end{figure}

\section{Cardiac Motion Scoring}

The proposed network Cardiac-MOS takes as input the myocardial segment sequences represented in the polar coordinate system. We first describe  how to convert the spatial cardiac MR sequences into the polar coordinate system, and then describe how Cardiac-MOS works.

\subsection{Myocardium segment re-sampling in polar coordinate system}
To ensure a uniform scoring procedure for all the 16 regions in spite of their position, and to leverage the powerful representation of neural network for motion information extraction, the fan-shaped regions obtained following the AHA 17-segment model of left ventricle~\cite{Lang2015Recommendations} (Fig.~\ref{fig_16seg_polar} (b)) are converted into the polar coordinate system. For segments of apical slice, we re-sample them along the angular dimension, therefore resulting in segments of the same size with those of basal and mid-slices (Fig.~\ref{fig_16seg_polar} (c)). Such conversion and re-sampling make it more convenient to capture the motion information and model the spatial dependencies since most motions fall into the same orientation. 
Denote the obtained $i$th ($i=1,...16$) segment sequence for patient $p$ as $X_{p,i}\in \mathcal{R}^{r,a,t}$, where $r, a$ are the sampling numbers along the radial and angular dimension, and $t$ is the frame number in the cardiac cycle. In this work, $r$ equals half of the image size, $a$ is 60 for all the segments.

\begin{figure}[!t]
	\centering
	\includegraphics[width=9.5cm]{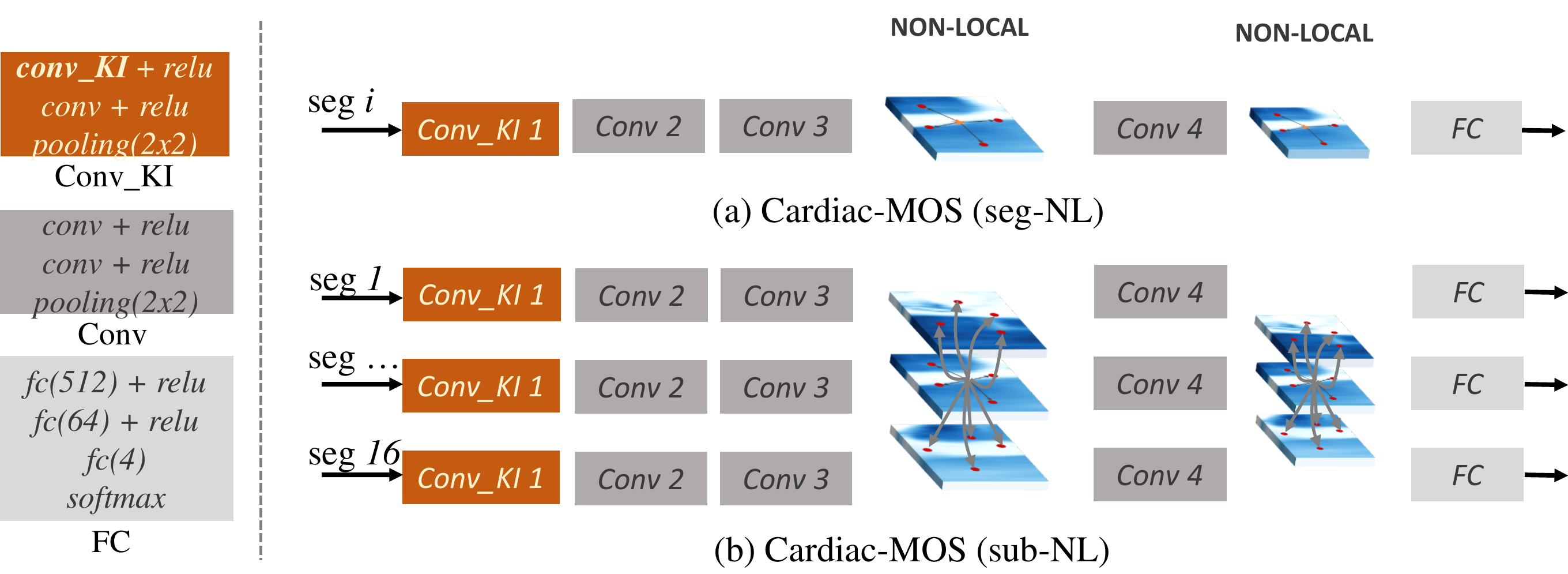}
	\caption{Our Cardiac-MOS models the distant non-local relationship of convolution features with Non-local neural networks in both segment-level ((a) Cardiac-MOS (seg-NL)) and subject-level ((b) Cardiac-MOS (sub-NL)). Kernel interpolation (conv-KI) is deployed to extract motion information from sequences of various lengths.}
	\label{fig_network}
\end{figure}

\subsection{Motion scoring neural network}
With the segment sequence $X_{p,i}$, the task of motion scoring aims to rate its motion function following the above mentioned scoring system $\mathbf{s}_{p,i}\in \{0,1,2,3\}$. The proposed Cardiac-MOS (Fig.~\ref{fig_network}) contains four successive convolution blocks to extract discriminative motion-aware information from each segment sequence. Specifically, to extract the motion information from sequences of various lengths, the first convolution block is equipped with a kernel-interpolation based convolution (conv-KI) layer, which is applied along the temporal dimension. To capture the subtle motion difference between segments, non-local relationship of local convolutional features is modeled with non-local neural network. Finally, a fully connection block with Softmax layer is used for motion scoring. 

\begin{figure}[!t]
	\centering
	\begin{minipage}[b]{0.4\textwidth}
		\includegraphics[width=\textwidth]{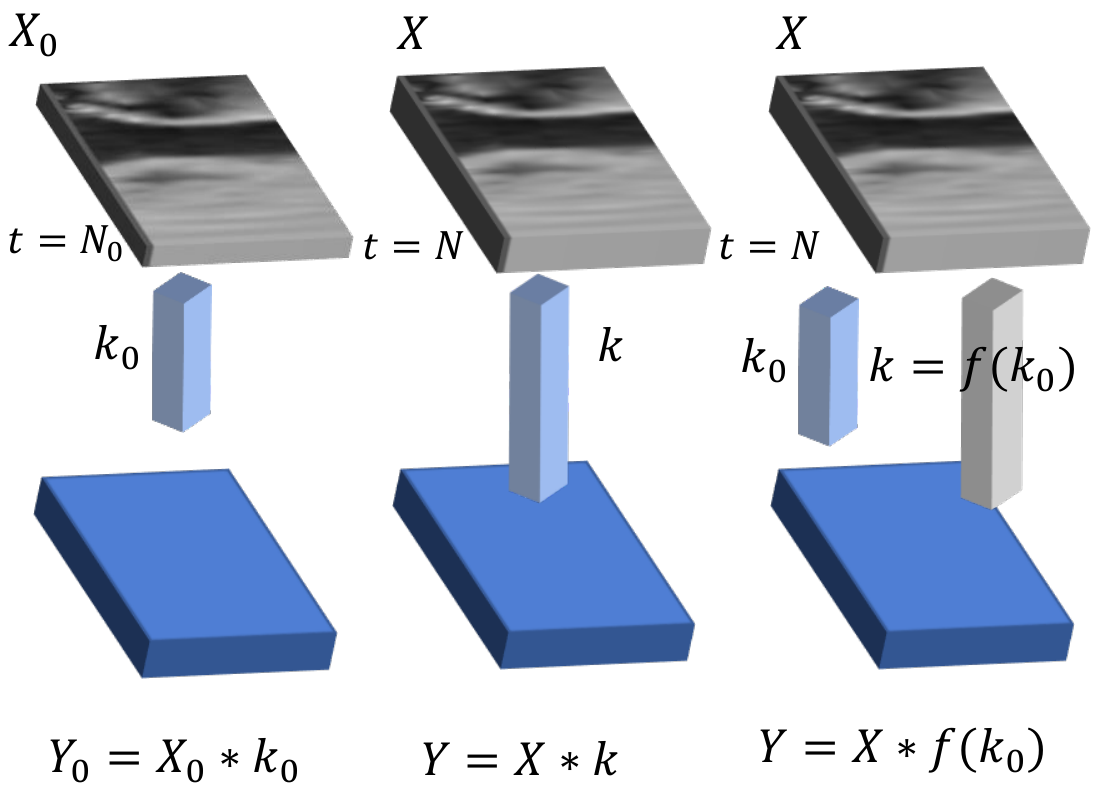}
		\caption{The proposed conv-KI layer extracts motion information from segment sequences of different lengths by interpolating $k_0$ to form the new kernel.}
		\label{fig_kernel_interp}
	\end{minipage}
	\hfill
	\begin{minipage}[b]{0.54\textwidth}
		\includegraphics[width=\textwidth]{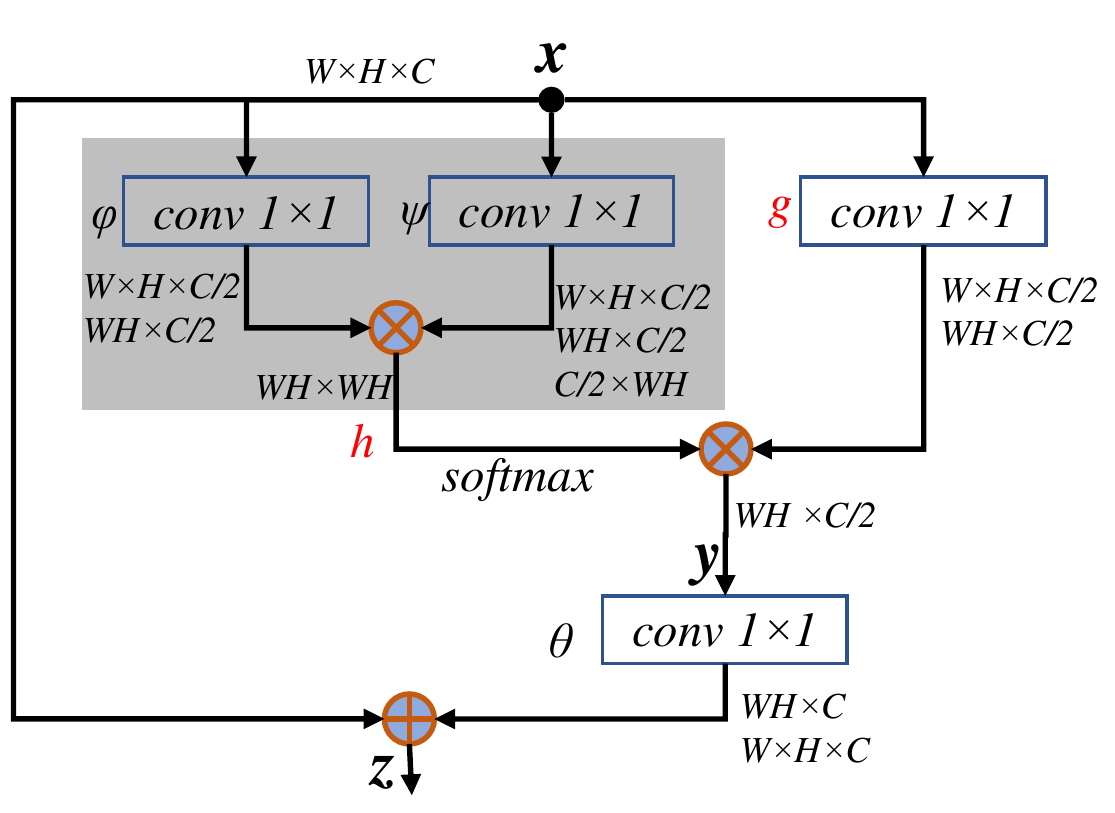}
		\caption{Non-Local block employed in Fig.~\ref{fig_network}, which explores relationships between distant non-local response in feature maps $\mathbf{x}$ of the convolution layer.}
		\label{fig_nonlocal}
	\end{minipage}
\end{figure}

\vspace{-3mm}
\subsubsection{Motion extraction with kernel-interpolation based convolution layer} 
For input sequence $X_{p,i}$ with frame number $t=N_0$, we first employ a $1\times 1\times t\times n_{o}$ convolution kernel $k_0$ with $n_o$ the filter number, to extract its motion information (Fig.~\ref{fig_kernel_interp} left). Each filter in $k_0$ acts as a template to match the temporal profile in every spatial position. 
However, such a fixed-shape kernel cannot be applied to the sequence with a different length $t=N$ (Fig.~\ref{fig_kernel_interp} middle). We propose a kernel-interpolation based convolution (conv-KI) layer to reuse $k_0$ (Fig.~\ref{fig_kernel_interp} right) when the sequence length changes. Denote $X$ and $Y$ the input and output of a convolution layer, then in a conv-KI layer, $Y=X*f(k_0)$, where $k=f(k_0)$ can be any parametric or nonparametric function that makes $k$ match the frame number of the new sequence. With conv-KI layer, the proposed network can be trained with and applied to cardiac sequences of multiple sources. In this work, bilinear interpolation is used for $f$.

\vspace{-3mm}
\subsubsection{Segment- and Subject-level Non-local modeling}
Cardiac motion score is determined by the motion patterns within one segment and its relative difference to other segments. As shown in Fig.~\ref{fig_16seg_polar} (c), patches with similar appearance exist within one segment (patches in blue) and across all segments (patches in red) of the subject. Only the convolution layer cannot model such relationship due to their local receptive field. To model such long-range relationships, non-local (NL) blocks are deployed in Cardiac-MOS, as shown in Fig.~\ref{fig_network}(a) and (b) for segment- and subject-level NL properties, respectively. 
NL operation for neural network~\cite{Wang2017nonlocal} was inspired by the non-local means method in computer vision. A general NL operation computes the response $\mathbf{y}$ at a position $i$ as weighted sum of the features at all positions:
\begin{equation}\label{eq_nl}
\mathbf{y}_i=\frac{1}{\mathcal{C}(\mathbf{x})}\sum_{\forall j}h(\mathbf{x}_i,\mathbf{x}_j)g(\mathbf{x}_j),
\end{equation}
where the pairwise function $h$ computes the similarity between two positions, $g$ computes a representation of the input feature, and $\mathcal{C}(\mathbf{x})=\sum_{\forall j}h(\mathbf{x}_i,\mathbf{x}_j)$ is a normalization factor. In this work, we use the embedded Gaussian~\cite{Wang2017nonlocal} for $h$:
\begin{equation}
h(\mathbf{x}_i,\mathbf{x}_j)=\exp^{\phi(\mathbf{x}_i)^T\psi(\mathbf{x}_j)},
\end{equation}
which makes $\frac{1}{\mathcal{C}(\mathbf{x})}h(\mathbf{x}_i,\mathbf{x}_j)$ as a softmax function. 
The computation of a NL block in neural network is shown in Fig.~\ref{fig_nonlocal}:
\begin{equation}
\mathbf{z}_i=\theta(\mathbf{y}_i)+\mathbf{x}_i,
\end{equation}
where the residual connection $+\mathbf{x}_i$ allows a NL block to be inserted into any pre-trained model, while keeping its initial behavior. $\phi$, $\psi$ and $\theta$ are implemented as convolution with $1\times 1$ kernel.

In this work, NL blocks are innovatively deployed into Cardiac-MOS in two different ways:
1) Cardiac-MOS (seg-NL), as shown in Fig.~\ref{fig_network} (a), where the position index $j$ traverse all possible positions in one segment, therefore segment-level non-local property is explored; 
and 2) Cardiac-MOS (sub-NL), as shown in Fig.~\ref{fig_network} (b), where $j$ traverse all possible positions of all segments in one batch, thus subject-level non-local property is explored. While seg-NL helps capture the motion features robustly for each segment, sub-NL helps capture the relative change of motion patterns between segments. During network training, one batch contains 16 segments, i.e., all segments from one subject. 


\section{Dataset and Configurations}\label{sec_dataset}
To validate the proposed Cardiac-MOS, cardiac MR sequences of 90 subjects (5775 images) are collected from scanners of multiple vendors, with spatial resolution of 0.6445$\sim$1.9792 mm/pixel, frame number of 20 (65 subjects) or 25 (25 subjects) in one cardiac cycle. Various pathologies are in presence, including dilated cardiomyopathy, ventricular hypertrophy, infarction, ischemia, scar formation, etc. For each subject, three representative short-axis cine sequences are chosen from the basal, mid, and apical left ventricle. Segment-wise ground truth of motion score is obtained from the radiologists' report and expert's visual inspection, resulting in a total of 1440 segments in the dataset. The distribution of motions score is \{794, 348, 207,  91\} for the four motion types $s_{p,i}$. 

For each sequence, three landmarks, i.e, the junctions of left and right ventricles, and the center point of cavity, are manually pointed to crop the left ventricle and to align its orientation. Normalization of the myocardium size is conducted by resizing the cropped images to $160\times 160$. Normalization of the intensity is conducted by contrast-limited adaptive histogram equalization.

Our network is implemented in Tensorflow, with cross-entropy loss used during training. Three-fold cross validation (split according to subject, not segment) is used to benchmark the performance. Following~\cite{Wang2017nonlocal}, we first train a baseline CNN network, where no NL-block is employed, and then finetune the two non-local networks from this pre-trained model.

\section{Results and Analysis}
The performance of Cardiac-MOS is benchmarked with the task of motion scoring, and motion score index (MSI, calculated as the average of the scores of all segments for each subject~\cite{Lang2015Recommendations}) estimation. Classification accuracy ($acc_{ms}$) and Pearson correlation $\rho_{msi}$ are used for evaluation. To compare with existing work on binary abnormality detection, classification accuracy ($acc_{ad}$) and Kappa value~\cite{viera2005understanding} ($\kappa_{ad}$, which is a more convincing index considering the prevalence of positive and negative samples) are also evaluated.       

As shown in Table~\ref{table_cardiac_mos}, the proposed Cardiac-MOS achieves effective performance, with correlation of 0.926 for MSI estimation, and accuracy of 0.774 for motions scoring. For reference, the only reported inter-observer correlation for MSI is 0.85~\cite{bjornstad1996interobserver} for echocardiography. Therefore, the proposed Cardiac-MOS has great potential in clinical practice of cardiac motion analysis. 
Besides, the non-local modeling effectively improves the performance of Cardiac-MOS upon the baseline CNN model, proving its capability of capturing subtle motion difference between segments.
Among the four configurations of NL blocks, Cardiac-MOS (sub-NL-1) performs better than the other configurations, revealing the importance of subject-level non-local modeling for cardiac motion scoring,
which explores the mutual dependencies of local features across all 16 segments, therefore is capable of capturing the subtle relative change of motion patterns between segments. This inter-segment dependencies are better modeled with higher level convolution features (with more context information), thus Cardiac-MOS (sub-NL-2) performs slightly inferior to Cardiac-MOS (sub-NL-1).
When compared with existing methods for abnormality detection (Table~\ref{table_methods_review}), the proposed Cardiac-MOS achieves the best Kappa value among all the competitors despite the unfair settings, proving the effectiveness of Cardiac-MOS.

\begin{table}[!t]
	\caption{Performance of Cardiac-MOS with various configuration of NL blocks. (seg-NL-1: one segment-level NL block after Conv4; seg-NL-2: two segment-level NL blocks after Conv3 and Conv4.)}
	\label{table_cardiac_mos}
	\centering
	\begin{tabular}{l|c|c|c|c}
		\hline
		&$acc_{ms}$(\%)&$\rho_{msi}$&$acc_{ad}$(\%)&$\kappa_{ad}$\\
		\hline
		baseline CNN&74.7&0.913&85.8&0.711\\
		Cardiac-MOS (seg-NL-1)&75.6&0.907&86.3&0.720\\
		Cardiac-MOS (seg-NL-2)&76.1&0.914&87.0&0.735\\
		Cardiac-MOS (sub-NL-1)&\textbf{77.4}&\textbf{0.926}&\textbf{87.8}&\textbf{0.749}\\
		Cardiac-MOS (sub-NL-2)&76.4&0.919&87.6&0.746\\
		\hline
	\end{tabular}
\end{table}

\section{Conclusions}
We proposed the first effective solution Cardiac-MOS for automated cardiac motion scoring from MR sequences. Cardiac-MOS is capable of extracting the complex motion information from cardiac sequences of various lengths with a kernel-interpolation based convolution layer, learning discriminative features with hierarchy convolution layers, and capturing subtle differences among motion patterns by modeling long-range spatial dependency with segment- and subject-level non-local networks. Performance of Cardiac-MOS on a MR dataset of 90 subjects demonstrated its great potential in future clinical application.       


\bibliographystyle{splncs03}
\bibliography{cardiacmotion}

\end{document}